\documentclass[conference]{IEEEtran}
\IEEEoverridecommandlockouts

\usepackage{cite}
\usepackage{amsmath,amssymb,amsfonts}
\usepackage{algorithmic}
\usepackage{graphicx}
\usepackage{textcomp}
\usepackage[table,xcdraw]{xcolor}  
\usepackage{subcaption}
\usepackage{hyperref} 
\usepackage{booktabs}
\usepackage{array}
\usepackage{caption}

\def\BibTeX{{\rm B\kern-.05em{\sc i\kern-.025em b}\kern-.08em
    T\kern-.1667em\lower.7ex\hbox{E}\kern-.125emX}}

\begin{document}

    \title{DriveBLIP2: Attention-Guided Explanation Generation for Complex Driving Scenarios}

\author{Shihong Ling$^{1}$, Yue Wan$^{1}$, Xiaowei Jia$^{1}$, and Na Du$^{1}$
\thanks{$^{1}$School of Computing and Information, University of Pittsburgh, Pittsburgh, PA, USA}%
}

\maketitle

\begin{abstract}
This paper introduces a new framework, DriveBLIP2, built upon the BLIP2-OPT architecture, to generate accurate and contextually relevant explanations for emerging driving scenarios. While existing vision-language models perform well in general tasks, they encounter difficulties in understanding complex, multi-object environments, particularly in real-time applications such as autonomous driving, where the rapid identification of key objects is crucial. To address this limitation, an Attention Map Generator is proposed to highlight significant objects relevant to driving decisions within critical video frames. By directing the model's focus to these key regions, the generated attention map helps produce clear and relevant explanations, enabling drivers to better understand the vehicle's decision-making process in critical situations. Evaluations on the DRAMA dataset reveal significant improvements in explanation quality, as indicated by higher BLEU, ROUGE, CIDEr, and SPICE scores compared to baseline models. These findings underscore the potential of targeted attention mechanisms in vision-language models for enhancing explainability in real-time autonomous driving.
\end{abstract}

\section{Introduction}
In recent years, large language models (LLMs) have made significant progress in natural language processing, powering a wide range of applications from machine translation to text generation. These models, such as GPT\cite{b1,b21} and BERT\cite{b2}, have demonstrated remarkable abilities to understand and generate human-like text for a wide range of tasks. 

Researchers further enhance these models into multi-modal LLMs, which integrate visual, auditory, and textual data to tackle complex real-world tasks such as image captioning, video understanding, and robotic control. 
Notably, vision-language models, such as CLIP\cite{b4}, ALIGN\cite{b5}, and Florence\cite{b6}, have emerged as powerful tools for bridging vision and language, leading to remarkable performance in applications like image captioning, visual question answering, and image retrieval.

As multi-modal LLMs continue to advance, their potential to improve the explainability of human-interactive systems has gained significant attention. In contexts such as autonomous driving, explainability is essential for building driver trust in the vehicle’s decision-making process. Without sufficient explainability, drivers may perceive autonomous vehicles as ``black boxes", leading to either under-reliance or over-reliance on the system during critical moments \cite{b7,b34}. This lack of clarity can result in delayed reactions or unnecessary interventions, highlighting the need for autonomous systems to generate clear and focused explanations. 

Multi-modal LLMs have shown their promise in generating explanations in driving scenarios~\cite{b8,b9,b10,b11,b12,b13,b14,b15}. For instance, Talk2Car\cite{b18} provides natural language explanations based on user commands, facilitating interaction between drivers and autonomous vehicles. Additionally, DriveGPT4\cite{b15} integrates vision and language to provide end-to-end interpretability in autonomous driving by leveraging visual inputs to generate text explanations for vehicle actions. Other models such as SurrealDriver\cite{b14} and GPT-Driver\cite{b13} demonstrate the growing capabilities of LLMs to handle complex driving scenarios by generating driving-related textual responses based on video input. 
Despite these advances, most of these models do not identify or prioritize key objects in a scene, such as crossing pedestrians and cutting-in vehicles. Instead, they generate explanations that are either too broad or overly detailed with less focus on relevant information essential for vehicle's driving decisions. Such explanations could  potentially overwhelm drivers, particularly in critical situations where concise and relevant information is needed.



To address this challenge, we propose a novel framework, DriveBLIP2, which 
incorporates an attention mechanism tailored specifically for autonomous driving and integrates it with the BLIP2-OPT  architecture~\cite{b16,b17}. In particular, our proposed method introduces an Attention Map Generator that identifies significant objects in key video frames, allowing the model to focus on the most critical regions of the scene. By incorporating such information into explanation generation, our method ensures that the generated explanations are more tailored to the identified objects, and thus can provide drivers with accurate and concise information 
during emergent driving scenarios. Through evaluations on real driving datasets, we demonstrate that the proposed method can effectively reduce irrelevant details and emphasize key elements, which contributes to  
clearer and more effective outputs suited to real-time autonomous driving applications. The superiority of explanations generated by the proposed method is also confirmed through extensive comparison with its vanilla end-to-end counterpart and several baselines.
\section{Related Work}
\subsection{Vision-Language Models for Autonomous Vehicle}
The incorporation of multi-modal LLMs into autonomous driving has shown significant promise in enhancing both vehicle perception and control systems. These models combine computer vision, human language, and sensor data (if applicable) to generate actionable insights and explanations in real-time driving scenarios. Recent developments include models like Driving with LLMs\cite{b8}, which use LLaMA\cite{b19} as a backbone to handle perception and control tasks through a vision-language interface. By leveraging a pre-trained LLM, this model processes both visual inputs and language commands to generate driving actions. Similarly, Talk2BEV\cite{b9} introduces a language-enhanced Bird's Eye View (BEV) map system, using pre-trained models like Flan5XXL\cite{b20} to interpret driving scenes and respond to user queries in natural language.

Another notable model is GAIA-1\cite{b10}, which focuses on planning by integrating visual and language inputs to generate realistic driving scenarios. This model uses a video-based prompt approach to enhance the understanding of road conditions and decision-making processes. LMaZP\cite{b11} and Dilu\cite{b12} extend the use of GPT-3 and GPT-4 in planning and control, using language-conditioned text inputs to generate action plans for driving. Models like SurrealDriver\cite{b14} and GPT-Driver\cite{b13} leverage the power of GPT-4 to interpret complex driving scenarios and generate textual explanations for trajectory planning and control actions.

Among these various multi-modal models, the architecture of BLIP2 \cite{b33} stands out due to its modular design, which separates the vision encoder, Q-Former, and LLM decoder. This structure allows for selective fine-tuning of different components and provides a flexible platform for external modifications, such as introducing task-specific attention mechanisms.

\subsection{Attention Mechanisms in Multi-modal Models}

Attention plays a pivotal role in the success of multi-modal models by enabling them to selectively focus on the most relevant parts of the input data.  In transformer-based architectures, attention mechanisms allow the model to weigh the importance of different input tokens, either from text, image, or other modalities, which significantly enhances the model's ability to handle multiple types of information.

Most vision-language models, such as CLIP\cite{b4} and ALIGN\cite{b5}, focus more on the global semantic similarity between the two modalities, while ignoring fine-grained alignments between visual objects and mentioned entities. While these models demonstrate impressive zero-shot capabilities and perform well on general tasks, they struggle to specifically focus on key objects within a scene. Recently, efforts have been made to introduce multi-modal grounding capabilities into LLMs\cite{b27,b28}. These models can handle general tasks like question answering and captioning, while also generating grounding information that links visual objects to corresponding entities. However, directly applying these methods to autonomous driving presents several challenges. First, rather than focusing on the alignment of explicitly mentioned objects, we are more concerned with inferring the implicit objects of significance from the perspective of ego vehicles. Second, the complexity of driving scenes involves dynamic and interacting objects, which requires not just static grounding, but continuous tracking and understanding of the evolving context in real time.
\vspace{-6pt}

\begin{figure*}[htbp]
    \centering
    \subcaptionbox{Main Model Architecture\label{fig:main_model}}{
        \includegraphics[width=0.9\textwidth]{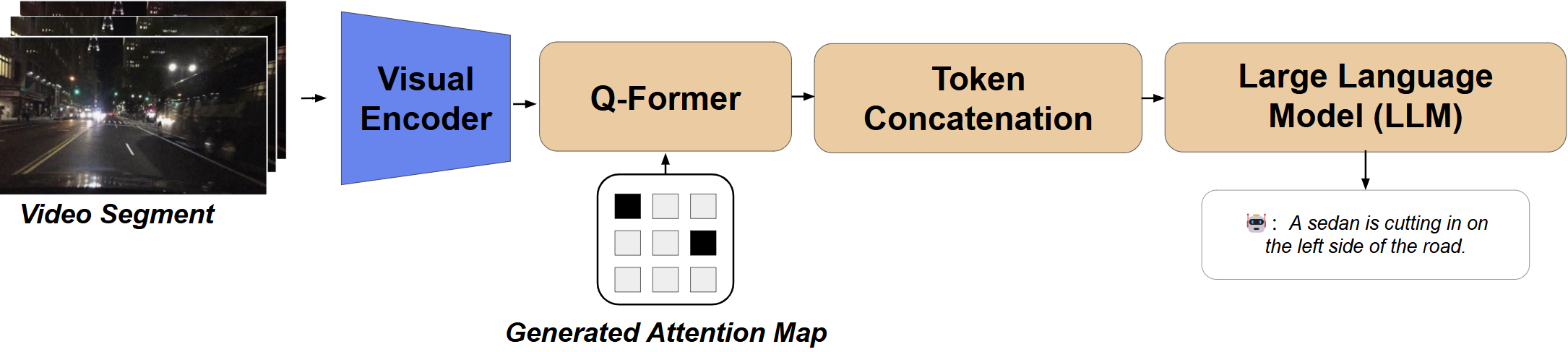}
    }
    
    \subcaptionbox{Attention Map Generator\label{fig:attention_map_generator}}{
        \includegraphics[width=0.9\textwidth]{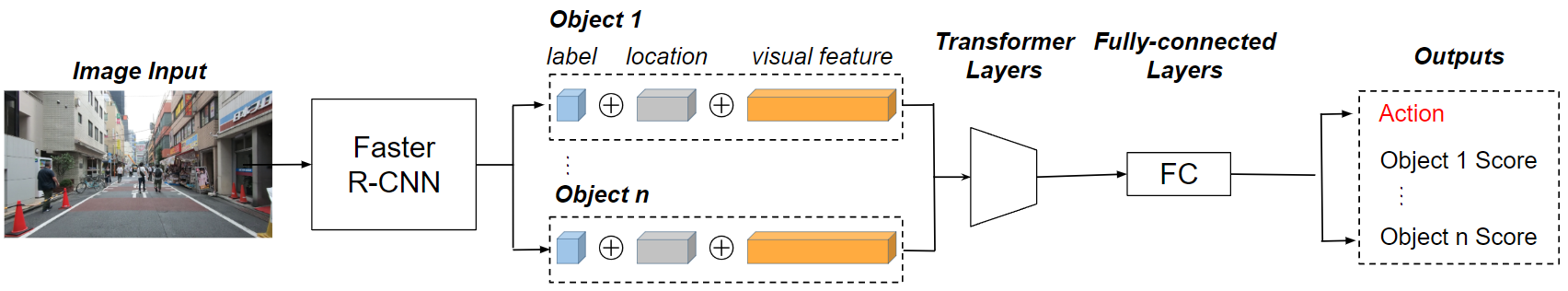}
    }

    \caption{Model architectures: (a) Main Model, (b) Attention Map Generator}
    \label{fig:model_architectures}
    \vspace{-6pt}
\end{figure*}
\section{Methodology}


In this section, we describe the framework of our proposed method, DriveBLIP2, which extends the BLIP2-OPT architecture \cite{b33} to handle video input and generate concise, contextually relevant explanations for emergent driving scenarios. The model first processes short video sequences by encoding each frame through an image encoder. An Attention Map Generator is then applied to dynamically identify and highlight the most important objects within the scene for driving. This map guides the model to focus on the key regions, ensuring that the generated explanations emphasize the most relevant elements. Once the attention map is computed, the model cross-references the highlighted regions with the visual features extracted from the frames. The explanations generated by the model are then constructed based on these key elements. The architectures of both the primary model and the Attention Map Generator are depicted in Figure \ref{fig:model_architectures}.

\subsection{Preliminaries}

\textbf{BLIP2}\cite{b33} is a state-of-the-art vision-language model that integrates visual data and natural language processing through a multi-stage architecture. The model consists of three main components: an image encoder, a Querying Transformer (Q-Former), and a Large Language Model (LLM). The BLIP2-OPT variant specifically utilizes the OPT language model, a transformer-based model designed for generating detailed and contextually accurate descriptions based on visual input.

In BLIP2, the process begins with the image encoder, which transforms an input image into a sequence of visual tokens. These tokens are then processed by the Q-Former, a transformer-based module that aligns the representation space between the visual encoder and the LLMs. The Q-Former employs a cross-attention mechanism that enables the LLM to focus on specific regions of the image, thereby facilitating the generation of text that describes the most relevant aspects of the visual scene for driving.

\subsection{Overall Model Structure}

\subsubsection{Temporal Token Concatenation}

While BLIP2 was initially designed to process static images or independent image sequences, our task requires handling short video clips. To achieve this, we modify the model to process sequences of frames by concatenating the visual tokens generated from each frame in a pre-defined temporal order after the Q-Former processes them. This approach ensures that the model captures the temporal progression of the scene, allowing the LLM to generate a contextually relevant narrative that reflects the dynamic nature of the video input.

Specifically, given a video \(V\) composed of frames \(\{I_1, I_2, \dots, I_n\}\), each frame \(I_i\) is first processed by the Transformer-based image encoder, which produces a sequence of visual features. These tokens are then processed by the Q-Former, resulting in a token sequence \(T_{I_i}\). The concatenated token sequence \(T_V\) is then represented as:
\vspace{=2mm}
\[
T_V = [T_{I_1}, T_{I_2}, \dots, T_{I_n}]
\]
This sequence is then treated as prompt tokens for the LLMs.
\vspace{=2mm}
\subsubsection{Object-of-significance Attention}

In the original BLIP2 model,  the receptive field of attention in Q-Former spans across all visual inputs. This means that the attention map used in the original model assigns similar weights to all patches at the beginning of training, which is not ideal for scenarios where certain objects, such as a car suddenly cutting into the lane, are more critical than others.

To address this, we propose an auxiliary model called the Attention Map Generator, which estimates the objects' significance in emergent driving scenes. The raw pixel-level attention map is then used to assign the key object score to all image patches that contain the corresponding object, quantifying the importance of each patch. These importance values are then fed into the Q-Former part of the primary model. The Q-Former is modified to utilize this generated attention map during the calculation of cross-attention. Specifically, it randomly masks out patches that are deemed non-significant based on the attention map, thereby reducing their influence on the following computations. Unlike the original model, which assigns equal importance to all the patches (i.e., equivalent to using an attention map with all values set to 1), the modified Q-Former now selectively ignores non-significant patches. After this refinement, the Q-Former extracts visual tokens from the processed frame representations. These visual tokens are then passed to the pre-trained LLM's decoder, which generates textual explanations describing the scene.

\subsection{Attention Map Generator}

\subsubsection{Attention Computation}

The \textbf{Attention Map Generator} plays a crucial role in determining which objects in a video frame should be considered significant. This module is based on a pre-trained Faster R-CNN model, which is widely recognized for its object detection capabilities. For each detected object \(o_i\) within a frame, we extract several key features: the object’s index \(I_i\), its bounding box coordinates \(B_i\) (the bounding box \(B_i\) typically consists of four coordinates that define the region of the image where the object is located), and the corresponding pixel values \(P_i\). These features are combined into a comprehensive feature vector \(F_i\) for each object $i$:
\vspace{=2mm}
\[
F_i = [I_i, B_i, P_i]
\]
\vspace{=2mm}
These feature vectors are processed through transformer layers to calculate attention weights for each object, reflecting their importance within the scene. Based on these attention weights, the model generates two outputs: the significance scores for each object and the predicted action of the ego-vehicle (e.g., braking or slowing down). By prioritizing objects based on their relevance to the driving action, the model can focus on critical elements that directly influence the vehicle's actions. By prioritizing objects based on their relevance to the driving task, the model can focus on essential elements that directly influence the vehicle's actions. For example, if a pedestrian suddenly crosses the road in front of the vehicle, the model assigns higher attention weights to the pedestrian, ensuring that the vehicle responds by yielding appropriately. 

To optimize the attention map, we apply two key loss functions: Intersection over Union (IoU) loss and Cross Entropy (CE) loss. The IoU loss encourages alignment of detected bounding boxes with the human-annotated critical objects, while the CE loss is used for multi-class classification of the ego-vehicle’s reaction. The total loss \(L\) is computed as:
\vspace{=2mm}
\[
L = L_{\text{IoU}} + L_{\text{CE}}
\]
\vspace{=2mm}
where
\vspace{=2mm}
\[
L_{\text{IoU}} = -\sum_{i} \text{IoU}(B_{\text{pred}, i}, B_{\text{GT}, i}) \times \log A_{\text{sig}, i}
\]
\vspace{=2mm}
Here, \(B_{\text{pred}, i}\) represents the predicted bounding box for the \(i\)-th object, \(B_{\text{GT}, i}\) represents the ground truth bounding box for the \(i\)-th object, and \(A_{\text{sig}, i}\) is the attention weight assigned to the \(i\)-th object.
\vspace{=2mm}
\[
L_{\text{CE}} = \text{CrossEntropy}(\text{action\_pred}, \text{action\_GT})
\]
\vspace{=2mm}
where \(\text{action\_pred}\) represents the predicted action or reaction of the ego-vehicle, and \(\text{action\_GT}\) represents the ground truth action or reaction of the ego-vehicle.

\subsubsection{Attention Scores Post-processing}

After the significance scores are computed for each object, the system clusters objects based on their scores. Objects with higher significance scores are grouped, and their bounding box coordinates are used to construct the Attention Map for Significant Objects. 

The patch-level attention map is generated by mapping the locations of the significant objects, based on their bounding box coordinates, to the corresponding image patches. 
The frame is divided into a grid of non-overlapping patches, each representing a fixed region of the frame. For each significant object, its bounding box is projected onto this grid, and any patch that is fully or partially covered by the bounding box is assigned a value of 1 in the attention map, indicating high significance. Patches that are not associated with any significant objects are assigned a value of 0.
This map will be further utilized by Q-Former as mentioned.

\section{Experiments and Results}

\subsection{Dataset}
We conduct experiments on the DRAMA dataset\cite{b22}, which provides 17,785 interactive driving scenarios, each captured in a 2-second video clip from urban Tokyo. The dataset highlights situations where the ego-vehicle perceives risk. Each video includes bounding boxes for significant objects in keyframes and explanations of perceived risks and object behaviors. Ego-vehicle reactions, such as stopping, are also annotated. Furthermore, to align with our model, we simplify the explanations into a template: \textit{significant object name + action/status + position}, removing redundancies for more concise and relevant outputs.

\subsection{Training Procedure}
The training process for our framework involves two key components: the main model based on the BLIP2 architecture and the Attention Map Generator, which enhances the model’s ability to focus on significant objects within the scene.

\subsubsection{Main Model Training}
The main model, based on BLIP2, is fine-tuned on language modeling tasks, where only the Querying Transformer (Q-Former) and the Large Language Model (LLM) components are optimized.


We use the Adam optimizer with a learning rate of \(1 \times 10^{-4}\) and a batch size of 32. The learning rate is adjusted using a step scheduler, with a step size of 50 and a decay rate of 0.1. The target modules for training include all parameters associated with the Q-Former and LLM. We train the model for 500 epochs, to minimize the language modeling Cross Entropy loss between the generated token sequence and the ground truth explanations.

\subsubsection{Attention Map Generator Training}
The goal of training the Attention Map Generator is to produce an attention map for significant objects that highlights critical patches in each video's keyframe. To improve the accuracy of the attention map, the model is trained to predict the ego-vehicle's reaction to emergent scenarios, establishing a connection between the significance of objects and the vehicle's decision-making process. We train the model using the Adam optimizer with a learning rate of \(1 \times 10^{-4}\) and a batch size of 16, aiming to minimize the combined IoU and Cross Entropy loss as introduced in the previous section.



\subsection{Evaluation Metrics}
We evaluate explanation quality using standard text generation metrics. First, we measure the validation language modeling loss to assess alignment with ground truth explanations. For text evaluation, we use BLEU\cite{b23}, ROUGE\cite{b24}, CIDEr\cite{b25}, and SPICE\cite{b26} metrics. BLEU and ROUGE focus on n-gram overlap, while CIDEr and SPICE are used to assess the semantic accuracy and content richness of the descriptions. These metrics are chosen to comprehensively evaluate both the surface-level accuracy (BLEU, ROUGE) and the deeper semantic quality (CIDEr, SPICE) of the generated explanations, ensuring that the model not only replicates the structure of reference explanations but also captures the key information. In addition, we evaluate the auxiliary model by measuring its ability to identify the critical object within a set of detected objects. We report the percentage of cases where the highest-scoring object matches the ground truth and a more relaxed criterion where the ground truth is among the top three highest-scoring objects.

\subsection{Results}
\subsubsection{Main Model Performance}
To assess the performance of our proposed framework, we evaluate three variants: 1) DriveBLIP2 without object-of-significance attention, 2) DriveBLIP2 with object-level attention map, and 3) DriveBLIP2 with predicted patch-level attention map, which corresponds to our main model. The first variant serves as our baseline. The second variant receives the ground truth information about significant objects, which serves as the performance upperbound of the object-of-significance guided explanations. In contrast, DriveBLIP2 with the inferred patch-level attention map incorporates our attention map generator, which identifies groups of significant objects rather than specific objects. 

As shown in Table \ref{tab:performance_comparison}, the introduction of object-based attention significantly improves the BLEU, ROUGE, CIDEr, and SPICE scores compared to the performance of DriveBLIP2 without attention guidance. This improvement demonstrates that incorporating object-specific attention into the model allows it to generate more accurate and meaningful descriptions, capturing the critical elements of the driving scenario more effectively. The DriveBLIP2 with patch-level attention map, which utilizes the learned attention map generator, achieves performance nearly on par with the object-based attention model. This result highlights the practical effectiveness of the attention map generator, as it can approximate significant object identification well enough to produce explanations comparable to those generated with direct object input. The inclusion of both the object-based and customized attention map models demonstrates the importance of guiding the model to focus on critical areas of the image. However, relying on manually annotated objects is impractical in real-time settings, so the attention map generator offers a more feasible solution.
\begin{table}[htbp]
\centering
\caption{Performance comparison of different DriveBLIP2 variants. - corresponds to the DriveBLIP2 without attention guidance. * indicates that the attention is predicted by the Attention Map Generator. }
\begin{tabular}{|c|c|c|c|c|c|}
\hline
\textbf{Attention} & \textbf{CE Loss} & \textbf{BLEU} & \textbf{ROUGE} & \textbf{CIDEr} & \textbf{SPICE} \\
\hline
- & 0.53 & 0.44 & 0.60 & 1.54 & 0.27 \\
Object-level & 0.39 & 0.52 & 0.74 & 2.16 & 0.36 \\
Patch-level* & 0.41 & 0.51 & 0.75 & 2.15 & 0.34 \\
\hline
\end{tabular}
\label{tab:performance_comparison}
\end{table}
Additionally, we compare our method with other popular methods for language generation from visual inputs: BLIP\cite{b31}, BLIP2\cite{b33}, VIT + GPT\cite{b32}, and VideoLLaMA\cite{b29,b30}. For BLIP, BLIP2, and VIT + GPT, the models are fine-tuned with the dataset on key video frames. In the case of VideoLLaMA, we use a well-designed prompt to guide the model in identifying the most important object in the emergent driving scenario from the driver’s perspective, and the description follows the same format as our annotated explanations: \textit{significant object name + action/status + position}. 
To enhance the evaluation of our method, we further compare it with other vision-language models specifically designed for autonomous vehicles: Context-VLM\cite{b36}, Drive-GPT\cite{b13}, and EM-VLM4AD\cite{b37}. Context-VLM and EM-VLM4AD are fine-tuned on our dataset according to their respective protocols to ensure a fair evaluation. Drive-GPT, on the other hand, leverages GPT-4-Turbo for visual-language understanding and employs a well-structured prompt, similar to VideoLLaMA.
As shown in Table \ref{tab:performance_comparison_2}, Our model outperforms others, achieving the highest scores across all metrics.
\begin{table}[htbp] 
\centering 
\caption{Performance Comparison of Different Models} 
\begin{tabular}{|c|c|c|c|c|} 
\hline 
\textbf{Model} & \textbf{BLEU} & \textbf{ROUGE} & \textbf{CIDEr} & \textbf{SPICE} \\ \hline 
BLIP & 0.21 & 0.52 & 1.30 & 0.27 \\ 
BLIP2 & 0.32 & 0.58 & 1.54 & 0.28 \\
VIT + GPT & 0.20 & 0.49 & 1.41 & 0.27 \\
VideoLLaMA & 0.40 & 0.59 & 1.63 & 0.26 \\ 
Context-VLM & 0.34 & 0.47 & 1.49 & 0.26 \\ 
Drive-GPT (gpt-4-turbo) & 0.43 & 0.37 & 1.52 & 0.27 \\ 
EM-VLM4AD & 0.47 & 0.69 & 1.92 & 0.30 \\ 
DriveBLIP2 (Ours) & 0.51 & 0.75 & 2.15 & 0.34 \\ 
\hline 
\end{tabular} 
\label{tab:performance_comparison_2} 
\vspace{-6pt}
\end{table}

\subsubsection{Attention Map Generator Performance}
The accuracy of our attention map generator is evaluated by measuring its ability to select the correct significant object within a group of detected important objects. When selecting only the highest-scoring object, the model correctly identifies the ground truth critical object in 47.0\% of cases. However, when considering the top three highest-scoring objects, this accuracy increases to 63.2\%, meaning that the generator includes the correct object in the group approximately two-thirds of the time. This demonstrates that, while the generator may not always identify the exact object of interest, it consistently highlights a group of significant objects that guide the model in producing accurate explanations. The results confirm that the attention map generator is effective at identifying a group of important objects, and even though the exact object might not always be included, the model still benefits from the guidance provided by the attention map. 
\begin{figure*}[htbp]
  \centering
  \includegraphics[width=0.9\textwidth]{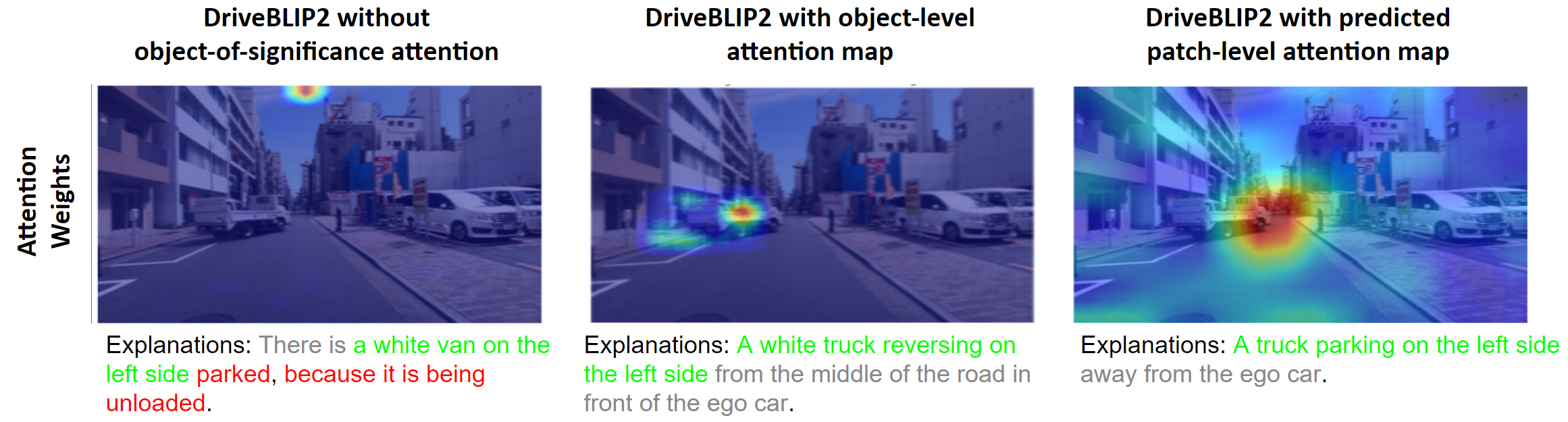}
    \vspace{-2mm}
  \caption{The attention map color scale ranges from red (strong focus) to blue (weak focus). In the generated explanations, green indicates correct information, red indicates incorrect information, and gray denotes unnecessary information.}
  \label{fig:compare-1}
\end{figure*}

\begin{figure*}[h]
  \centering
  \includegraphics[width=0.9\textwidth]{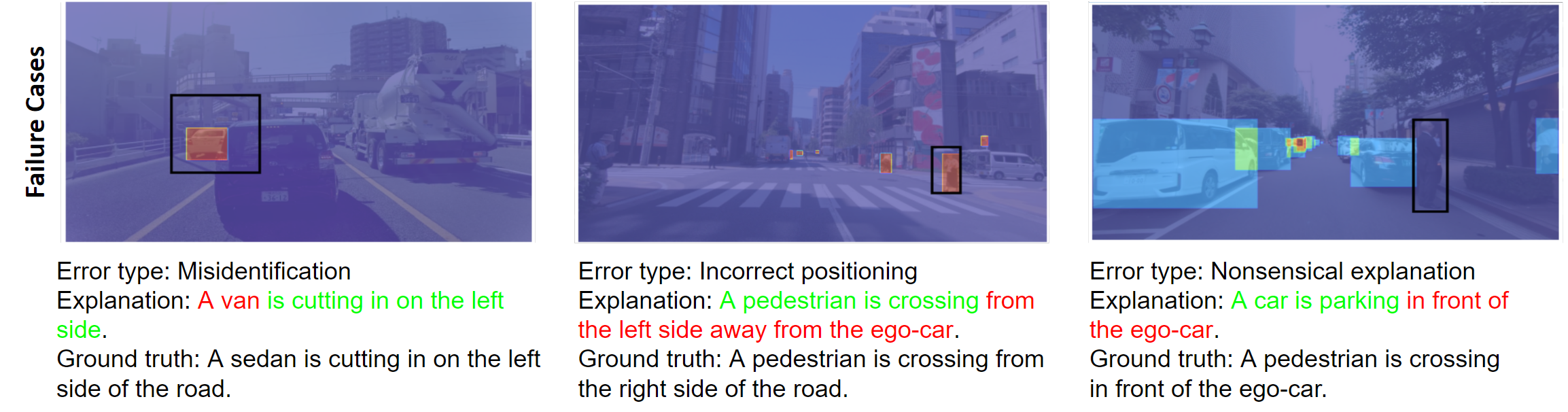}
    \vspace{-2mm}
  \caption{In the attention map, red indicates strong focus, blue indicates weak focus, and a black bounding box highlights the annotated significant object; in the generated explanations, green marks correct information, and red marks incorrect.}
  \label{fig:compare-2}
  \vspace{-6pt}
\end{figure*}

\section{Discussion}
Our results demonstrate that the attention mechanism introduced into the model significantly improves the quality of the generated explanations. By comparing the three variants of DriveBLIP2 as well as other models such as BLIP, VIT + GPT, VideoLLaMA, Context-VLM, Drive-GPT, and EM-VLM4AD, we observe notable differences in performance, as summarized in Tables \ref{tab:performance_comparison} and \ref{tab:performance_comparison_2}. The DriveBLIP2 with object-level attention map consistently achieves higher scores across BLEU, ROUGE, CIDEr, and SPICE, showing that explicit object information enables the model to generate more precise and informative explanations. 

DriveBLIP2 with patch-level attention map, which is the main model used for inference, performs comparably well, yielding results that closely mirror the object-based attention variant. This demonstrates that even when the model is only provided with approximate groups of significant objects, it can still produce explanations that are comparable in quality to those generated using manually annotated object locations. This finding supports the practicality of our approach in real-time applications, where manual annotation is not possible. 

To further analyze the effectiveness of our method, we visualize the object-of-significance attention and the corresponding generated text in Figure \ref{fig:compare-1}. These visualizations highlight the final attention weights during explanation generation, which are concentrated on key objects in the image. The attention mechanism allows the model to focus on the most relevant objects, and we observe a clear connection between these objects and the generated text. For example, in a scenario where the model is tasked with describing a pedestrian crossing the road, the attention is directed toward the pedestrian, and the generated text accurately reflects the pedestrian's actions and position. This indicates that the proposed attention mechanism enhances the model’s ability to generate explanations that align with the key parts of the scene.

Despite these improvements, there are several limitations to our approach as shown in Figure \ref{fig:compare-2}. The first issue we encounter is the misidentification of object labels. In some cases, the model incorrectly labels the boxed objects in its generated explanation, likely due to limitations in the pre-trained image encoder or biases within the dataset. For example, a few cars, which are labeled as ``sedan" instead of general ``car" in the ground truth explanations, are often mislabeled. This may be due to the insufficient exposure to such objects during training. Another issue arises in the positioning of objects within the generated explanations. In some instances, the model produces text that incorrectly describes the position of an object. One possible explanation for this error is that the model is overfitting to the positional bias of objects within the training data. This might lead the model to learn general patterns of where important objects typically appear, rather than developing a precise understanding of object positioning. Additionally, the fact that the ego-vehicle is not visible in the frames complicates the model's ability to determine the relative positions of objects. Since the camera is mounted on the ego-vehicle, the absence of a visible reference point makes it challenging for the model to infer positions relative to the vehicle. A third challenge involves the generation of nonsensical explanations, observed in high object density scenarios. In scenes containing too many objects, the model occasionally produces explanations that are misaligned from the actual visual content. While the cause of this issue is not entirely clear, the presence of many objects appears to increase the likelihood of such errors. This suggests that the model may have difficulty dealing with multiple objects simultaneously.

\section{Future Work}
Future work can focus on enhancing the framework to address its current limitations. Specifically, object recognition can be improved by expanding the dataset to include more diverse and complex driving scenarios. To handle complex multi-object scenes more effectively, the Attention Map Generator will be further refined to process object features with additional information, such as object status and its relationship to the ego-vehicle. Lastly, future research can explore using alternative large language models beyond the original OPT to assess whether other models can enhance the alignment between visual and textual data and improve the coherence of the generated explanations.

\section{Conclusion}
In this work, we address the challenge of generating contextually accurate explanations for emergent driving scenarios by extending the BLIP2-OPT architecture. We propose a novel framework, DriveBLIP2, which incorporates temporal information into the model and integrates the object-of-significance attention, driven by an Attention Map Generator, to guide the explanation generation by focusing on the most critical objects within a scene. 
Experimental results show that the introduction of object-based attention improves the quality of the generated explanations. With the patch-level attention map predicted by Attention Map Generator, DriveBLIP2 achieves a performance comparable to direct object-based attention, validating the practicality of our approach in real-world applications. Despite these successes, limitations remain in terms of object recognition in complex scenes with excessive objects. Addressing these limitations and improving the accuracy in selecting significant objects will be the focus of future work.


\end{document}